\documentclass[runningheads]{llncs}
\usepackage{graphicx}
\usepackage{amsmath}
\begin{document}
\title{Explainable Online Lane Change Predictions on a Digital Twin with a Layer Normalized LSTM and Layer-wise Relevance Propagation\thanks{This research was co-funded by the Bavarian Ministry of Economic Affairs, Regional Development and Energy, project Dependable AI, IBM Deutschland GmbH, and IBM Research, and was carried out within the Center for AI jointly founded by IBM and fortiss.}
}
\titlerunning{Explainable Online Lane Change Predictions on a Digital Twin}
%
\author{Christoph Wehner\inst{1}\orcidID{0000-0003-0421-4113} \and
Francis Powlesland\inst{2}\orcidID{0000-0002-8663-9109} \and
Bashar Altakrouri\inst{2}\orcidID{0000-0002-1157-6246} \and
Ute Schmid\inst{1,3}\orcidID{0000-0002-1301-0326}}
\authorrunning{C.~Wehner et al.}
%

\institute{University of Bamberg, Bamberg, Germany \\ \email{\{christoph.wehner, ute.schmid\}@uni-bamberg.de} \and
IBM Deutschland GmbH, Munich, Germany \\ 
\email{francis.powlesland1@ibm.com}, \email{bashar.tak@gmail.com}\\ \and
fortiss GmbH, Munich, Germany}

\maketitle              
\begin{abstract}

Artificial Intelligence and Digital Twins play an integral role in driving innovation in the domain of intelligent driving. Long short-term memory (LSTM) is a leading driver in the field of lane change prediction for manoeuvre anticipation. However, the decision-making process of such models is complex and non-transparent, hence reducing the trustworthiness of the smart solution. This work presents an innovative approach and a technical implementation for explaining lane change predictions of layer normalized LSTMs using Layer-wise Relevance Propagation (LRP). The core implementation includes consuming live data from a digital twin on a German highway, live predictions and explanations of lane changes by extending LRP to layer normalized LSTMs, and an interface for communicating and explaining the predictions to a human user. We aim to demonstrate faithful, understandable, and adaptable explanations of lane change prediction to increase the adoption and trustworthiness of AI systems that involve humans. Our research also emphases that explainability and state-of-the-art performance of ML models for manoeuvre anticipation go hand in hand without negatively affecting predictive effectiveness.

\end{abstract}

\section{Introduction}

Digital transformation trends such as Artificial Intelligence (AI), Digital Twins and internet of things plays an increasing and integral role in driving innovation and becoming ubiquitous in various domains and applications such as intelligent driving \cite{Khan19, Thevendran21}. These trends enable smart systems with novel capabilities that were never possible without AI. However, with the increasing adoption of these complex AI driven systems, new challenges emerge, especially related to the acceptance and trustworthiness of these systems by human users. Hence, there are increasing voices that demand more transparent and explainable AI models and systems. 

AI engines for predicting lane changes can be implemented using white-box models that come with the advantage of being explainable by default. Alternatively, black-box classifiers currently outperform white-box models on the lane change prediction task but struggle as they are not explainable \cite{xing19}. For mission critical tasks that involve human, such as predicting lane changes, both explainability and performance are crucial and equally important. 

This paper explores how the decisions making process of a complex black-box classifier can be made explicit and explained to a user. We present a state-of-the-art approach for lane change predictions that are explainable and a novel technical proof-of-concept implementation.  

\section{Related Work} 
Predicting lane changes for vehicles is an ongoing field of research. In their paper, Xing et al. \cite{xing19} presented a general discussion and a survey of the latest technology trends around this topic. 

The literature suggests various approaches to predict lane changing behaviour that mainly varying in the data and AI architecture used. Chen et al. \cite{chen19} aimed to train an attention-based deep reinforcement agent based on visual data in a simulated environment that predicts lane changes. Another approach, by Tang et al.  \cite{Tang19}, uses tabular data to train an adaptive fuzzy neural network to predict if a lane change takes place soon. Furthermore, a popular machine learning architecture for this type of problem is the recurrent neural network, as it is optimized to deal with problems related to time sequence analyses \cite{Patel18}. 

With the increasing attention on trustworthiness and transparency of machine learning models and systems, we are seeing a focus in recent literature on explainable models for the lane change prediction task, to be able to explain the reasons behind a predicted vehicle lane change for a human user or road stakeholder. The main goal is to move away from predicting with black-box models and aiming to increase the performance of white-box models like expert systems and other explainable classifiers \cite{Gallitz19, Gallitz20}. One paper by Dank et al. reformulated the prediction task based on tabular data to a regression problem \cite{furnkranz17}.

While white-box models come with the advantage of being explainable \cite{schwalbe21}, they are outperformed on the lane change prediction task by black-box classifiers \cite{xing19}. Nonetheless, the latter are not explainable. For safety critical tasks, both explainability and performance are crucial and equally important.

AI systems cannot be implemented without reliable data resources. Advancement in the area of Internet of Things (IoT) and Digital Twins within the automotive area, especially around autonomous and intelligent driving, can be seen in recent literature and successfully deployed projects and systems \cite{steyn2021development, kumar2018novel, el2020roads}. 
Alongside the vehicles themselves, infrastructure, such as roads and highways have also undergone modernization in places, so that these elements also can relay their "state" back to operators in real time. A real example of this is the Providentia++ Digital Twin \cite{kraemmer2019providentia}, which covers a section of Autobahn between Munich and Munich Airport. Here, the Providentia++ team decided to use cameras placed at regular points along the road, combined with visual recognition to identify vehicles. The setup is capable of relaying the position of every vehicle on the track, with a high level of accuracy and frequent update cycles. 

This paper shows an approach to combining data from the Providentia++ Digital Twin with an explainable machine learning model to predict lane changes in real-time. 
The following section shall introduce this approach. 

\section{Approach}
In this section, we present our suggested approach towards explainable lane change prediction supported by an extensible technical implementation. 

\subsection{Lane Change Predictions by a Layer Normalized LSTM}
The lane change predictions are computed by a layer normalized long short-term memory proposed for this purpose by Patel et al. in 2018 \cite{Patel18}. This section shall introduce the prediction model and its input features regarding relevant perspectives for generating explanations of its predictions. Please consult Patel et al.'s paper \cite{Patel18} for further information and evaluation of the ML model.

First, a layer normalized LSTM \cite{ba16} considers at each time step \(t\) a 1-dimensional array of vehicles \([v^i_t| \ \forall i \in [0,1,2,3,4,5,q]]\). The vehicles \(v^i\) with \(i \in [0,1,2,3,4,5]\) are the closest existing neighbours of \(v^q\). 
Each vehicle \(v^i\) at the time \(k\) is represented by the following array of features: 

\begin{equation}
v^i_k = [v_{x_k}^i, v_{y_k}^i, \psi_k^i, x_k^i, y_k^i, n_l^i, n_r^i],
\end{equation}

where \(x_k^i\) and \(y_k^i\) are the absolute world-fixed positions in meters, \(v_{x_k}^i\), \(v_{y_k}^i\) the respective velocities in meters per second, \(\psi_k^i\) is the heading angle of the vehicle in radiance and \(n_l^i\), \(n_r^i\) the number of lanes to the left and right.  
Furthermore, at each time step, the layer normalized LSTM considers as an input the previous cell state \(c_{k-0.5s}\) and the previous recurrent state \(h_{k-0.5s}\) \cite{hochreiter97}. Formally, the layer normalized LSTM is defined in Equation \ref{eqn:lstm} \cite{Patel18}.

\begin{equation} \label{eqn:lstm}
(h_{k},c_{k}) = lnLSTM([v_{k}^{0};...;v_{k}^{5};v_{k}^{q}],h_{k-1},c_{k-1})
\end{equation}

Its output at each time step is the cell state \(c_{k}\) and the recurrent state \(h_{k}\). For each prediction \(k \in [t_{-1.5s},t_{-1.0s},t_{-0.5s},t]\) time steps are shown to the layer normalized LSTM layer. \(t\) is the time the prediction is generated. Therefore, the layer normalized LSTM observes four frames of a vehicle and its surroundings within 1.5 seconds, before it creates a prediction. 

Layer normalization \cite{ba16} is applied before the non-linearities of the LSTM to increase its robustness. 

Layer normalization \(\vartheta(\cdot)\) is defined as follows:  

\begin{equation}
    \vartheta(a) = f \left[ \frac{g}{\sigma} \odot  (a - \mu) + b \right]   \quad
    \sigma = \sqrt{\frac{1}{H} \sum_{i = 1}^{H} ( a_{i} - \mu)^{2}}  \quad
    \mu = \frac{1}{H} \sum_{i = 1}^{H} a_{i}, 
    \label{eqn:normalizationlayer}
\end{equation}

where \( \mu \) is the mean of \( a \). \( a \) is the activation vector along the feature axis before the non-linearities of the gated interactions inside an LSTM cell. \( H \) 
denotes the number of hidden units in a layer, \( \sigma \) is the standard deviation of \( a \), \(g\) are the learned gain parameters, and \(b\) is a learned bias \cite{ba16}.

Ba et al. showed that layer normalization stabilizes the gradient \cite{ba16}. 
This results in a more stable and faster convergence of the validation loss to an optimum at training time and increases classification performance at inference time. 

The model's output represents if \(v^q\) changes to the left or right lane or stays on the same lane within the next 2.5 seconds. The labels are a one-hot encoding of the three classes. 

\subsection{Explanations of the Prediction Generated by LRP}
Layer normalized LSTM's show state-of-the-art performance at the lane change prediction task \cite{Patel18, xing19}. However, their decision-making process is considered a black box, as it is too complex and complicated to be understood by a user. 

We follow the increasing demand and research efforts to explain the decision-making process of a black-box classifier. 

The core of our proposed prediction engine applies the Layer-wise Relevance Propagation (LRP) attribution method on the layer normalized LSTM. We aim to make the decision-making process of the lane change prediction explicit, by identifying which part of the input is relevant for the classification. 

LRP assigns each input dimension of the layer normalized LSTM a relevance value. The relevance values represent how much each input dimension contributed to the prediction. 

LRP starts at the output layer, where the relevance for each neuron is set to be the prediction function value of the class to be explained \(f_c(x)\).
Layer by layer, the relevance is completely redistributed, from higher layer neurons to lower layer neurons by employing layer-specific LRP rules, where neurons that contribute most to the higher layer receive the most relevance from it, as explained in \cite{Bach15}. 

Arras et al. propose a chaining of (1) the LRP-\(\epsilon\) rule for the linear mappings, (2) the LRP-\(all\) rule for the gated interactions, and (3) the LRP accumulation rule to explain the interactions of a standard LSTM \cite{Arras2019}.

We extend their approach to layer normalized LSTM's by applying in addition 
the LRP-\(\Omega\) rule to the model-specific interaction of a layer normalized LSTM. In particular, we propose the novel LRP-\(\Omega\) rule to explain layer normalization. 

\begin{figure}[t]
\centering
\includegraphics[width=0.8\textwidth]{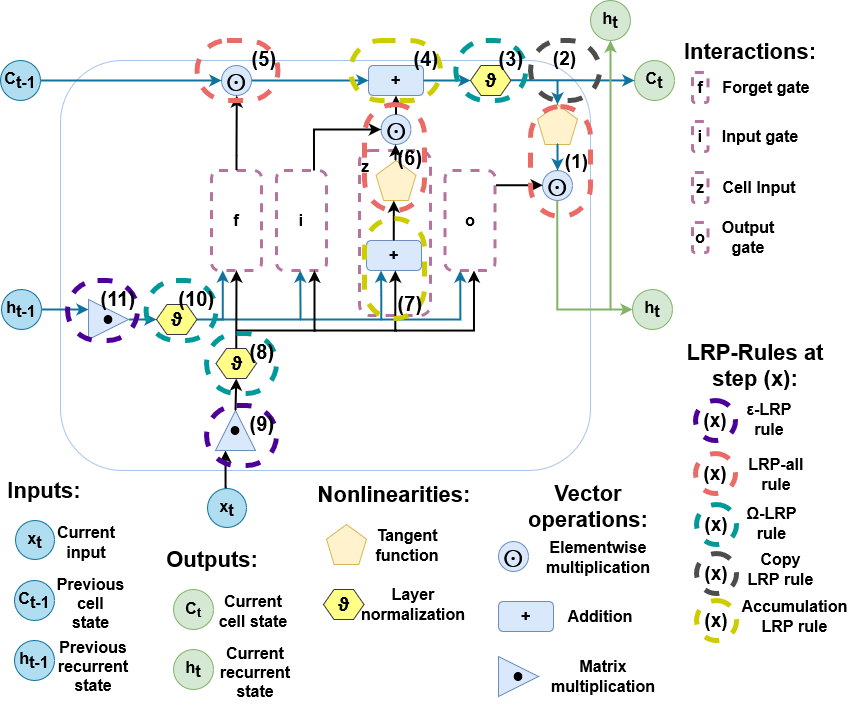}
\caption{Step by step chaining of the LPR rules for layer normalized LSTM's.} \label{fig:lrp_for_lnlstm}
\end{figure}

Figure \ref{fig:lrp_for_lnlstm} visualizes the LRP rule chaining for the layer normalized LSTM architecture. 

\emph{LRP-\(all\) Rule for the Gated Interactions} 
The relevance flow of the gated interactions in step (1), (5), and (6) of Figure \ref{fig:lrp_for_lnlstm} are retraced by the LRP-\(all\) rule.
With the LRP-\(all\) rule, all relevance 
flows to the source units. 
However, the gate units 
receive no relevance, as they do not hold information themselves but control the information flow \cite{Arras2019}.

\emph{LRP Accumulation Rule}
Accumulations are interactions appearing in step (4) and (7) on Figure \ref{fig:lrp_for_lnlstm}.
At accumulations, the relevance is split proportional to the magnitude of each addend, as suggested by 
\cite{ArrasMMS17}.

\emph{LRP-\(\epsilon\) Rule for Linear Mappings.}
Linear mappings are the interactions depicted in step (2)\footnote{Step (2) in Figure \ref{fig:lrp_for_lnlstm} is called the copy LRP rule. The copy LRP rule is a particular case of the LRP-\(\epsilon\) Rule, where one lower-level node and \(n\) upper-layer nodes exist, the weights are set to one, the bias is zero, and the activation function is linear.}, (9) and (11) of Figure \ref{fig:lrp_for_lnlstm}.
As suggested by \cite{Arras2019}, the LRP-\(\epsilon\) rule 
is used to retrace the relevance flow of the linear mapping. The linear mapping is equivalent to a dense layer with a linear activation function and a zero bias. 

\emph{LRP-\(\Omega\) Rule for Layer Normalization}
Layer normalization requires a specific LRP rule. According to our knowledge, LRP for layer normalization is not yet explored by the literature. In principle, layer normalization is similar to batch normalization \cite{IoffeS15}. However, they differ in the normalization dimension. While batch normalization normalizes over the whole batch, layer normalization normalizes over one instance \cite{IoffeS15, ba16}. We have explored and applied different LRP approaches for batch normalization to layer normalization, including the LRP identity rule\cite{Arras2019}, LRP-\(|z|\) \cite{Hui19}, LRP-\(\epsilon\) \cite{Hui19}, LRP fusion \cite{Guillemot20}, and LRP heuristic rule \cite{Alber19}.

\begin{figure}[t]
\centering
\includegraphics[width=0.45\textwidth]{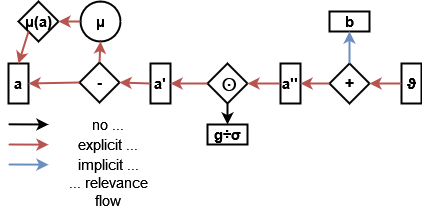}
\caption{Heuristic relevance decomposition of layer normalization in Equation \ref{eqn:normalizationlayer} by the LRP-\(\Omega\) rule. A black arrow signalizes no relevance flows to the term according to the LRP-\(\Omega\) rule. A red arrow signalizes relevance flows to the term, and the LRP-\(\Omega\) rule explicitly calculates it. Finally, a blue arrow signalizes relevance flows to the term, but it is not explicitly calculated in the LRP-\(\Omega\) rule as the term is a relevance sink.} \label{fig:lrp_for_ln}
\end{figure}

While the previous approaches made tremendous progress on explaining batch normalization, none fully consider the mean's impact on the relevance. Thus, we propose the novel LRP-\(\Omega\) rule for layer normalization. The LRP-\(\Omega\) rule decomposes layer normalization into a series of summations and scalings. This is shown in Figure \ref{fig:lrp_for_ln}. In particular, the LRP-\(\Omega\) rule acknowledges the impact of the mean shift in Equation \ref{eqn:normalizationlayer} on the relevance flow. Furthermore, it propagates the relevance assigned to the mean further down to the input of the layer normalization. 
The LRP-\(\Omega\) rule is formalized in Equation \ref{eqn:lrpnorm}. 

\begin{equation} 
    R_{i\leftarrow j} = (z_{i} - \frac{z_{i}}{H}) \cdot \frac{g_{i}}
    {\sigma} \cdot \frac{ R_j }{z_{j}} 
\label{eqn:lrpnorm}
\end{equation}

\(R\) is the relevance signal from the input value \(z\) of the higher layer unit \(j\) to the input value of the layer normalization unit \(i\). \(H\) is the length of the input to the layer normalization.

We outlined in this section how LRP calculates relevance values as explanations. The following section shall introduce how the relevance values are transformed into comprehensible explanations. 

\subsection{Comprehensible Explanations} \label{subsec:comprehensible_explanations}
LRP assigns relevance values to the \(4\times 49\) dimensions of the layer normalized LSTM's input. This is LRP's explanation of the model's prediction. 
The user of the maneuver anticipation system, i.e. the driver, will find 196 relevance values to be incomprehensible. Hence, we have adopted an aggregation approach that utilizes the adaptive nature of LRP in terms of dimensionality reduction. Due to LRPs' relevance conservation and redistribution property \cite{Bach15}, relevance values of terminal units can be added together without invalidating them. The aggregated value represents the relevance of the aggregated units. Thus, features in the input space can be aggregated to meaningful virtual super-features, and their aggregated attribution values represents their relevance for the classification.

The aggregation of relevance values makes it possible to communicate comprehensible explanations to the user. To explain the lane change prediction of the layer normalized LSTM, the time dimensions are aggregated. Therefore, the relevance values of input features representing a vehicle are added together over every time step: 

\begin{equation} 
    \forall i : R_{v^i} =  \sum_{\forall k} R_{v^i_k}
\end{equation}
Furthermore, the relevance values of the individual features of the vehicles are aggregated to the virtual and weighted super-features movement \(m\) and position \(p\):

\begin{gather}
    \forall i : R_{v^i} =  [R_{m^i}, R_{p^i}]; \nonumber \\ 
    with \  R_{m^i} = \frac{R_{v_{x}^i} + R_{v_{y}^i} + R_{\psi^i}}{3} 
           \ and \ R_{p^i} = \frac{R_{x_k^i} + R_{y_k^i} + R_{n_l^i} + R_{n_r^i}}{4}
\end{gather}

Finally, the three most relevant super-features are communicated to the user via the demonstrator in real-time, as shown in Figure \ref{fig:driver_perspective}.
The three most relevant features are visualized via their name and logo to the driver. In addition, a color scheme describes their relative impact on the classification. 

This section described how high dimensional explanations by LRP are reduced to make them comprehensible by the driver while steering a vehicle. Up next, the implementation details of the prototype and an evaluation shall be provided.

\subsection{Prototype architecture}
This prototype has been designed as a distributed set of containers, and as such can be deployed on any Kubernetes cluster with minimal configuration. 
This approach was chosen in order to maximize resiliency and redundancy across the application, whilst also logically separating concerns, permitting independent horizontal scaling. Up next, the elements that describe this prototype are detailed. 

\emph{Live Adaptor}
The Live Adaptor takes the protobuf stream from the digital twin, decodes and enriches it, so it can be consumed by other parts of the application. This optional step improved the workflow for the rest of the application by propagating the data as JSON. It also checks each vehicle that comes through the digital twin and assigns a UUID. This was necessary since the digital twin itself only assigns vehicle ID's in the cycle 1-10,000, meaning that we lose vehicle uniqueness if we record data that contains over 10,000 vehicles directly from the digital twin. To address this, we looked at each original vehicle ID coming through the digital twin, and checked to see when the ID was last present, if the original vehicle ID has not been present for a period of time, we assume that the vehicle is new, and it is issued with a new UUID. To store UUIDs we use redis as the in-memory cache, preserving state across application restarts and failures. Vehicle IDs contain no identifying information about the vehicle itself. 

\emph{Prediction Engine}
The prediction engine is composed of the \textbf{Prediction Model} and the \textbf{Service Broker}, that together enable the user to consume live predictions on demand, in a scalable way. To realize this, we kept the prediction model in a python container that communicates with the service broker over a standard HTTP protocol. Instead of having the prediction model handle connections to the user, we created a "sessioning" platform in the service broker which listens for user requests to open a "prediction session" for a specific UUID. While the session is open, snapshots are repeatedly collected from the live adaptor and are then sent off for inference. The novel element here is that the service broker can handle many connections at once, enabling multiple users, and handles all internal state about user sessions. Because of this, the prediction model itself is stateless and can be scaled horizontally. Once a vehicle leaves the digital twin, the service broker will automatically terminate the session, running any garbage collection.

\emph{General Considerations}
Because of the nature of this domain, specifically our data source being a live digital-twin, considerations were made across every facet of this project to make sure we utilised an event driven architecture. In practice, this meant heavily utilising technologies such as websockets for two-way communication between the system and the user, as well as using websockets to manage state across the system itself. Kubernetes was chosen as our platform as it allowed us to deploy highly customised containers with relative ease.

The architecture of the prototype is fully mapped. Up next, the explanations of the lane change predictions shall be evaluated and the prototype's GUI shall be discussed.

\section{Evaluation and Discussion}
This section evaluates the explanations provided by LRP in terms of their faithfulness to the layer normalized LSTM's behavior. Furthermore, the GUI of the prototype is presented and critically discussed. Please consult \cite{Patel18} for an in-depth evaluation and comparison of the layer normalized LSTM in contrast to other machine learning models for predicting lane changes.

\subsection{Evaluation of the Explanations}
A perturbation test is deployed to evaluate the explanation. 
The perturbation test is a behaviouristic approach to evaluate the faithfulness of an explanation. It asks if the explanations reflect the model's behaviour. 

For the perturbation test, classifications and attributions, i.e. relevance values, of a representative amount of instances are calculated. Next, the instances are split into correct and wrong classified instances. For the correct classified instances, the most important super-feature is occluded. The occluded instances are classified, and the accuracy is measured. Again, the remaining most relevant super-feature is occluded, and the model's accuracy on the newly created instances is measured. The previous step may be repeated until there is no more feature to occlude.
A faithful explanation method produces results that decrease the accuracy significantly more than randomly occluding features. 

To set the faithfulness of LRP for layer normalized LSTM's into context, LRP is compared with the attribution method Integrated Gradients \cite{Mukund2017}.
Furthermore, two versions of LRP are compared: LRP with the LRP identity rule \cite{Arras2019} applied to layer normalization and LRP with the LRP-\(\Omega\) rule for layer normalization. 

\begin{figure}[t]
\centering
\includegraphics[width=0.55\textwidth]{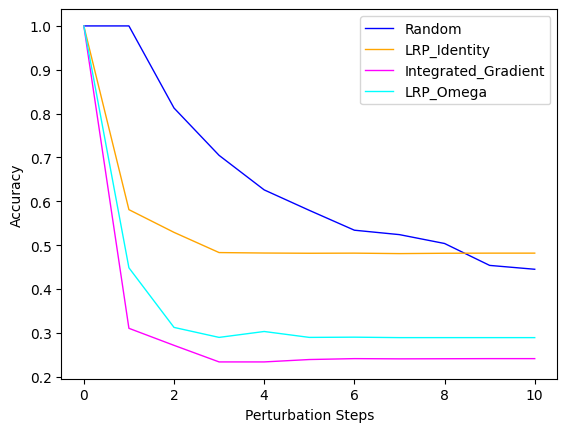}
\caption{Perturbation test on the layer normalized LSTM for the lane change prediction task. The perturbation test is conducted on 2315 instances. The instances are randomly drawn from a set, uniformly distributed over the labels.} \label{fig:perturbation_test}
\end{figure}

The results of the perturbation test are depicted in Figure \ref{fig:perturbation_test}. 

LRP with the LRP identity rule applied to layer normalization performs worst in the perturbation test. The perturbation test converges towards 50\% accuracy for this rule combination. After nine perturbation steps, it is outperformed by random occlusion. The rule captures the most relevant features accurately, but fails to distribute relevance to minor impactful parts of the input. 

The LRP-\(\Omega\) rule outperforms the LRP identity rule significantly. Heuristically redistributing the relevance from the layer's output to its input, while fully considering the impact of the mean on the relevance flow, increases the performance. The LRP-\(\Omega\) rule allows capturing the impact of every part of the input accurately.

Integrated Gradients outperforms LRP for the layer normalized LSTM in terms of faithfulness. At first glance, this is surprising. Gradient-based attribution methods tend to not perform well on standard LSTM's \cite{ArrasMMS17} because the gradient of the sigmoid- and tanh- non-linearities of the LSTM cell is close to zero for activations outside the interval [-4; 4] and respectively [-2; 2]. However, through the layer normalization, the inputs to the non-linearities are brought closer to those intervals, stabilizing the gradient and leading to faithful explanations of gradient-based attribution methods.

\emph{On the Computational Expenses of LRP}
We implemented Integrated Gradients and LRP in Tensorflow 2.4, running on a workstation with two Nvidia 2080TI, CUDA 11.1, 64GB RAM, and an AMD Ryzen Threadripper 2920X. LRP computed the explanations on average 10.47 times faster than Integrated Gradients for 2335 randomly drawn instances. This is due to approximating integrals being computationally expensive. 

LRP performs in terms of faithfulness comparably to Integrated Gradients while being significantly more computationally efficient. Thus, LRP is our method of choice for explaining the online lane changes predictions. 

Next, the GUI, where the prediction and explanation by LRP are presented, is discussed. 

\subsection{Discussion of the Prototype's GUI}
\begin{figure}[t]
\centering
\includegraphics[width=0.6\textwidth]{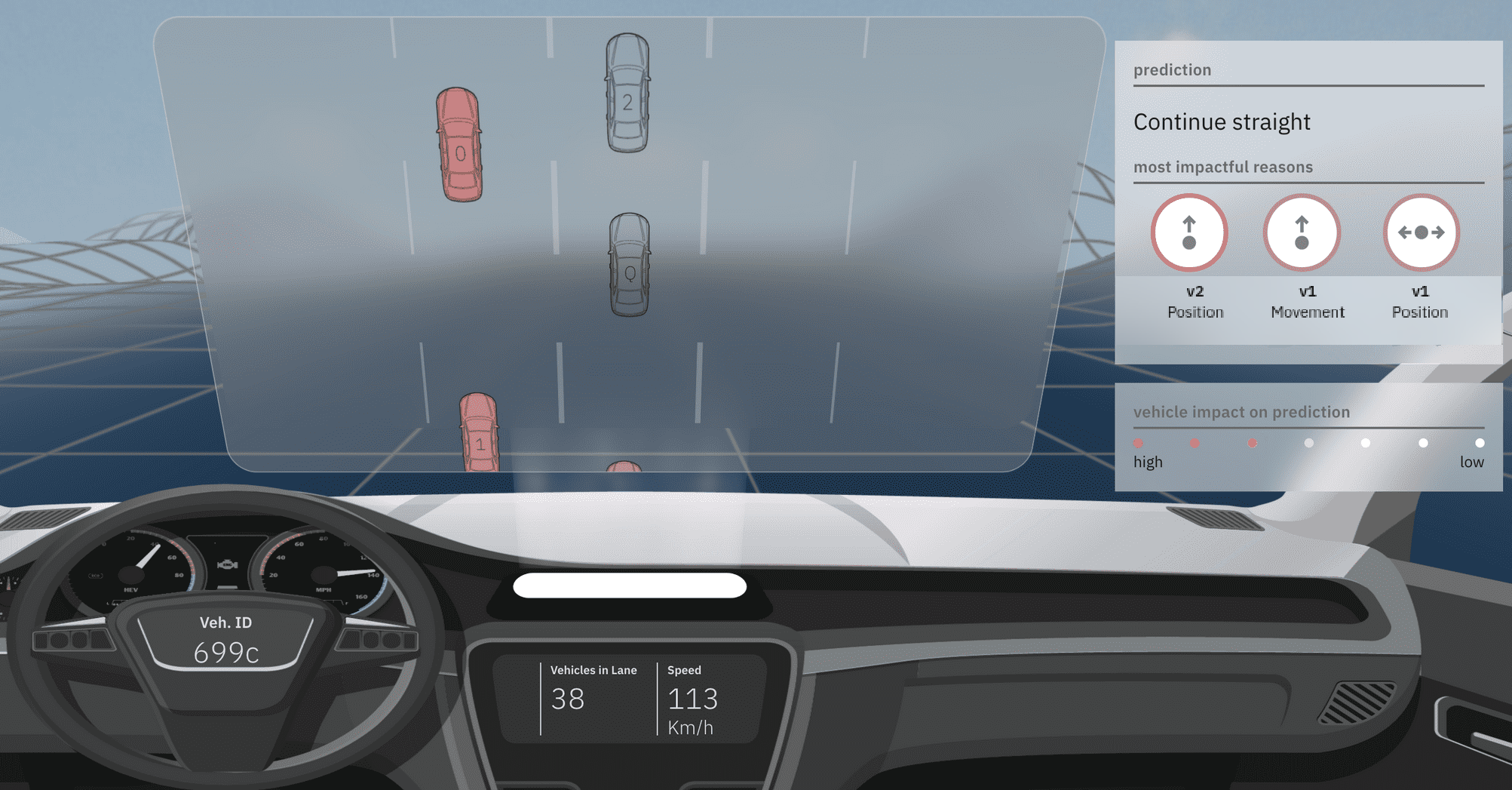}
\caption{The "driver perspective", showing our selected vehicle (highlighted with "Q"), the surrounding vehicles, and their impact on the latest prediction.} \label{fig:driver_perspective}
\end{figure}

The visualization component is the user facing web application that shows the capabilities of the demonstrator (Figure \ref{fig:driver_perspective}). This allows the user to "jump in" to a vehicle and get various insights as if they were driving the vehicle themselves. The user can see real-time stats such as the nearest neighbours, number of vehicles on the road, the next prediction and the reasons associated with it. The UI also instructs the prediction engine to start or stop predictions for a specific vehicle, rendering the output. In the explainability domain, our chosen approach here is to use a heatmap, where the neighbouring vehicles change color depending upon their actions and how much they impacted the latest prediction. 

\section{Conclusion}
This paper showcased how to predict and explain lane changes given live data of a digital twin. For that reason, the layer normalized LSTM is outlined as a state-of-art prediction model. However, its decision-making process is too complicated and complex to be understood by a user. In safety-critical applications like lane change predictions, a user must understand the reasoning of the prediction engine. Thus, we developed LRP for layer normalized LSTM's to make the decision-making process of the layer normalized LSTM explicit. LRP shows performant results in terms of faithfulness to the models' behavior while being computationally lightweight. Thus, it is the method of choice for explaining the lane change predictions in real-time. 
Furthermore, this paper gave implementation insides on how to realize a scalable, high-performance prototype for making explainable lane change predictions. In addition, we presented the user interface and critically discussed it. 
Future work includes implementing multimodal communication strategies of the computed relevance values beyond heatmaps. And the communication strategies of the prototype shall be evaluated in a user study.
Furthermore, it is an interesting open question on how to use the explanation of the model's prediction so that the user interactively improves the prediction model to make it more performant and trustful. 

Attribution methods provide deep insights into a black box machine learning model's decision-making process. Let us use those insights to create more trustful and safe machine learning applications. 
%
%
%
\bibliographystyle{splncs04}
%
\bibliography{53}

\end{document}